\documentclass[runningheads]{llncs}
\usepackage{graphicx}
\usepackage{orcidlink}

\usepackage{algorithm}
\usepackage{algcompatible}
\usepackage{algpseudocode}

\usepackage{amsmath,amssymb}
\usepackage{cleveref}
\usepackage{xfrac}
\usepackage{subcaption}
\usepackage{stmaryrd}
\usepackage{booktabs}
\usepackage{dirtytalk}
\usepackage{multirow}
\usepackage{adjustbox}
\usepackage{bm}
\usepackage{tabularx}

\begin{document}
\title{Zero-Shot Anomaly Detection with Pre-trained Segmentation Models}
%
%
\author{Matthew Baugh\inst{1}\orcidlink{0000-0001-6252-7658} \and
James Batten\inst{1}\orcidlink{0000-0002-8028-5709} \and
Johanna P. M\"uller\inst{2}\orcidlink{0000-0001-8636-7986} \and
Bernhard Kainz\inst{1,2}\orcidlink{0000-0002-7813-5023}}
\authorrunning{M. Baugh et al.}
%
\institute{Imperial College London, UK \\
\email{matthew.baugh17@imperial.ac.uk}
\and Friedrich--Alexander University Erlangen--N\"urnberg, DE}
\maketitle              
\begin{abstract}
This technical report outlines our submission to the zero-shot track of the Visual Anomaly and Novelty Detection (VAND) 2023 Challenge. 
Building on the performance of the WinCLIP framework, we aim to enhance the system's localization capabilities by integrating zero-shot segmentation models.
In addition, we perform foreground instance segmentation which enables the model to focus on the relevant parts of the image, thus allowing the models to better identify small or subtle deviations.
Our pipeline requires no external data or information, allowing for it to be directly applied to new datasets.
Our team (Variance Vigilance Vanguard) ranked third in the zero-shot track of the VAND challenge, and achieve an average F1-max score of 81.5/24.2 at a sample/pixel level on the VisA dataset.
\end{abstract}

\section{Introduction}

The creation of dedicated datasets for industrial anomaly detection has led to a heightened interest in the area, and with it huge progress.
Because of this, the state of the art has advanced so far that unsupervised industrial anomaly detection appears close to solved, with new methods often reporting an Area
Under the Receiver Operating Characteristic (AUROC) above 98.0 at both a sample and pixel level \cite{Liu_2023_CVPR,Roth_2022_CVPR,Tien_2023_CVPR} on the MVTec dataset \cite{bergmann2021mvtec}.
However, for these methods to perform well they often require a high number of normal samples to train on, making such algorithms difficult to apply in a real-world setting.
There is therefore a need for more data-efficient methods, able to identify anomalies when trained on only a few normal samples.  
WinCLIP\cite{Jeong2023} pushed this idea further, proposing a zero-shot method anomaly detection and localisation, leveraging language guidance to provide a signal for normality in the absence of normal images.

We adapt WinCLIP by incorporating zero-shot segmentation models to better localise anomalies.
We also use foreground segmentation to focus our model on each instance of the object within the image, leading to better segmenting of smaller anomalies.
Through this, our method achieves an average F1-max score of 24.2 at a pixel-level on the VisA dataset\cite{spotdiff_visa} and ranked third on the zero-shot track of the Visual Anomaly and Novelty Detection (VAND) 2023 Challenge.

\section{Method}

\begin{figure}
    \centering
    \includegraphics[width=\linewidth]{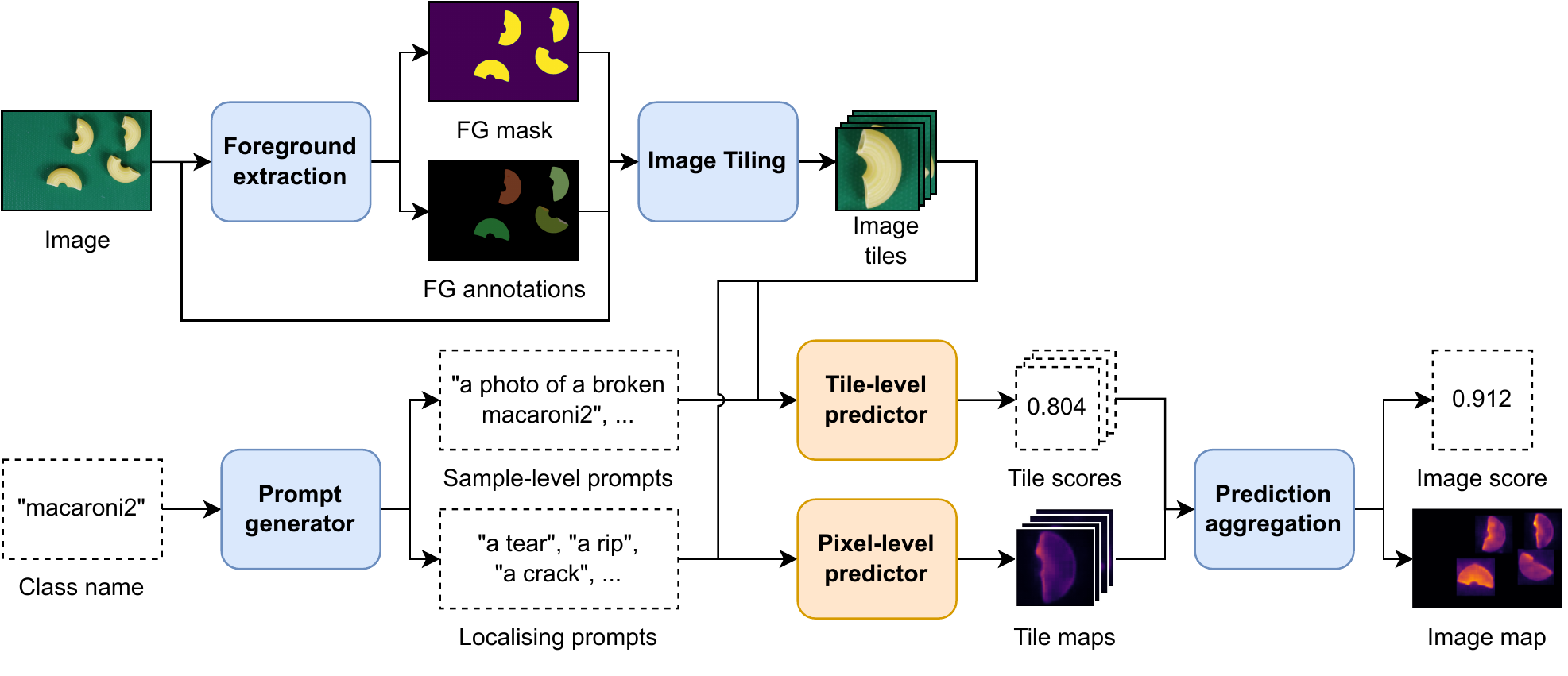}
    \caption{Our zero-shot anomaly detection pipeline.}
    \label{fig:pipeline}
\end{figure}

Our method maintains the core principles of WinCLIP \cite{Jeong2023} by using CLIP-based models to identify anomalous examples, but in order to better localise the anomalies we use a combination of zero-shot segmentation models.
Our pipeline (Fig.~\ref{fig:pipeline}) can be broken down into foreground extraction, image tiling, prompt generation, prediction (at both a tile and pixel level) and finally prediction aggregation.

\noindent\textbf{Foreground extraction: }
To identify the foreground of the image we combine dichotomous image segmentation \cite{qin2022} with SegmentAnything (SAM)~\cite{kirillov2023segany}, using the dichotomous segmentation to filter the annotations produced by SAM to identify which form the foreground of the image.
These annotations are then combined to produce the final foreground mask.
We found that this performs better than directly using the dichotomous segmentation as it often overestimates the size of an object to include part of the shadows.
As SAM normally produced a separate annotation for the object and its shadow, our simple filtering (requiring 80\% of the mask to be covered by the dichotomous segmentation) mitigated most of these cases.

\noindent\textbf{Image tiling: }
The primary purpose of our image tiling was to divide the image up into each instance of the object.
To do this we crop a square centred on each connected component of the foreground mask with a minimum resolution of $352\times 352$, which is the maximum input size of our later models, as we found that using tiles smaller than this degraded performance as the images would be distorted as part of the model's preprocessing steps.
An exception to this was made for components with a higher aspect ratio ($>1.5$ in either axis) which SAM identified as having many constituent parts ($> 20$).
For these objects taking a square bounding box would mean that the background would take up a high proportion of the image, making it difficult for the segmentation models to notice deviations in the components of the object.
In these cases we tile along the long axis of the bounding box of the object, taking steps of half the short axis length, to ensure that all parts of the object are included in the centre of at least one tile.

\noindent\textbf{Prompt generator: }
We used WinCLIP's \cite{Jeong2023} compositional prompt ensemble for the sample-level prompts, only extending the list of normal and abnormal states in order to increase robustness (Fig.~\ref{fig:comp_prompt} a, b).
For the localising prompts we use a list of generic nouns that describe an anomalous region of an object (Fig.~\ref{fig:comp_prompt} c).
Such prompts were more suited than directly reusing the WinCLIP prompts to localise the defects, as the WinCLIP prompts describe the entire image so would produce much broader segmentations than those just describing the anomaly.
The same list of localising prompts is used for all classes.

\noindent\textbf{Tile-level predictor: }
To compute an anomaly score for each tile we take a similar approach to WinCLIP \cite{Jeong2023}, comparing the CLIP embeddings of the different sample-level prompts to the embedding of each image tile.
However, rather than averaging themselves, we average the alignment between each prompt and the tile image embedding.
We favour this as although you would expect the embeddings of the ``normal" prompts to form a single cluster, as they are describing the same concept of a normal object, the anomalous prompts are likely to be spread a lot more sparsely as being anomalous covers a wide variety of concepts.
This means that taking the average embedding of the anomalous prompts is not as meaningful, but comparing the average cosine similarities avoids this issue.

\noindent\textbf{Pixel-level predictor: }
For pixel-level predictions, for each tile we use CLIPSeg \cite{luddecke2022image} to produce a segmentation from each of the localising prompts.
We then use a harmonic average across the prompt segmentations to focus on the regions which consistently give a higher activation.

\noindent\textbf{Prediction aggregation: }
As the tile-level predictor is generally more accurate and robust than the pixel-level predictor, we scale the pixel-wise predictions for each tile by its corresponding tile-level prediction.
These tiles are then rearranged into the original image space, with pixels belonging to multiple tiles being averaged across the predictions.
To compute the sample-level prediction, we first average the tile-level predictions across each foreground component, as different tiles of the same foreground are often heterogeneous so have different distributions of tile-level scores.
We then take the average of the top 25\% of foreground component scores, which is particularly useful in the multi-instance cases as it avoids some of the noise present in the scoring of the normal foreground regions.

\section{Results}

Following WinCLIP~\cite{Jeong2023} we used the F1-score at the optimal threshold (F1-max) to assess our method as it is less influenced by class imbalance, which is particularly prevalent in the segmentation evaluation as the anomalies are often quite small relative to the size of the image.
We provide our F1-max at both a sample level (Tab.~\ref{tab:sample_results}) and pixel level (Tab.~\ref{tab:pixel_results}) on the VisA dataset~\cite{spotdiff_visa}, comparing to WinCLIP~\cite{Jeong2023} as a baseline while also including the results of the VAND challenge winner APRIL-GAN~\cite{chen2023zero}.
Our use of additional sample-level prompts and averaging the cosine similarities achieves a new state-of-the-art for sample-wise F1-max (81.5), while our pixel-wise performance greatly improves over the baseline.

\begin{table}[]
    \centering
    \resizebox{\columnwidth}{!}{%
    \begin{tabular}{lrrrrrrrrrrrrr}
    \toprule
     & pcb1 & pcb2 & pcb3 & pcb4 & capsules & candle & macaroni1 & macaroni2 & cashew & chewinggum & fryum & pipe\_fryum & Mean \\
    \midrule
    WinCLIP                     & 71.0 & 67.1 & \textbf{71.0} & 74.9 & 83.9 & \textbf{89.4 }& 74.2 & 69.8 & \textbf{88.4} & \textbf{94.8} & 82.7 & 80.7 & 79.0 \\
    APRIL-GAN                   & 66.9 & \textbf{70.1} & 66.7 & \textbf{87.3} & 77.6 & 77.8 & 71.1 & 69.1 & 84.8 & 93.7 & \textbf{91.7} & 87.7 & 78.7 \\
    Variance Vigilance Vanguard & \textbf{74.3} & 67.1 & 70.2 & \textbf{87.3} & \textbf{84.9} & 82.1 & \textbf{83.3} & \textbf{76.9} & 82.3 & 94.4 & 84.8 & \textbf{90.0} & \textbf{81.5} \\
    \bottomrule
    \end{tabular}%
    }
    \caption{Sample-wise results, F1-max compared with baseline WinCLIP and challenge winner APRIL-GAN\cite{chen2023zero}}
    \label{tab:sample_results}
\end{table}

\begin{table}[]
    \centering
    \resizebox{\columnwidth}{!}{%
    \begin{tabular}{lrrrrrrrrrrrrr}
    \toprule
     & pcb1 & pcb2 & pcb3 & pcb4 & capsules & candle & macaroni1 & macaroni2 & cashew & chewinggum & fryum & pipe\_fryum & Mean \\
    \midrule
    WinCLIP & 2.4 & 4.7 & 10.3 & 32.0 & 9.2 & 22.5 & 7.0 & 1.0 & 13.2 & 41.1 & 22.1 & 12.3 & 14.8 \\
    APRIL-GAN & 12.5 & 23.4 & 21.7 & 31.3 & 48.5 & 39.4 & 35.5 & 13.7 & 22.9 & 78.5 & 29.7 & 30.4 & 32.3 \\
    Variance Vigilance Vanguard & 29.5 & 11.0 & 4.7 & 21.7 & 31.9 & 20.2 & 24.6 & 7.2 & 24.5 & 63.4 & 31.3 & 19.6 & 24.2 \\
    \bottomrule
    \end{tabular}%
    }
    \caption{Pixel-wise results, F1-max compared with baseline WinCLIP and challenge winner APRIL-GAN\cite{chen2023zero}}
    \label{tab:pixel_results}
\end{table}

\section{Conclusion}

We have greatly improved the segmentation ability of WinCLIP by incorporating zero-shot segmentation models.
However, there is certainly more scope for improvement in the localising ability, as many of the models we use struggle due to the magnitude of the domain shift from their original testing data to that of industrial anomaly detection.
This problem was amplified by many of the anomalies being exceedingly small and subtle.
As foundation models continue to progress we are excited to see how their better representations can be leveraged to better solve the task of zero-shot anomaly detection.
At a sample level, our results improve incrementally over WinCLIP~\cite{Jeong2023}, but there is still much work to be done to elevate zero-shot anomaly detection to be closer to the performance of unsupervised models.

%
%
%
\bibliographystyle{splncs04}
\bibliography{mybibliography}

\newpage

\section{Appendix}

\begin{figure*}
\noindent\begin{minipage}[t]{0.32\linewidth}
(a) Additional normal state-level prompts

{\tt \small
\begin{itemize}
    \item "good [o]"
    \item "normal [o]"
    \item "amazing [o]"
    \item "pristine [o]"
    \item "undamaged [o]"
    \item "[o] in good condition"
    \item "unbroken [o]"
    \item "[o] without any imperfections"
    \item "[o] without any scratches"
    \item "[o] without any marks"
    \item "complete [o]"
    \item "new [o]"
\end{itemize}
}

\end{minipage}
\hfill
\begin{minipage}[t]{0.33\linewidth}
(b) Additional anomalous state-level prompts

{\tt \small
\begin{itemize}
    \item "broken [o]"
    \item "bad [o]"
    \item "flawed [o]"
    \item "defective [o]"
    \item "[o] in poor condition"
    \item "worn [o]"
    \item "[o] with scratches"
    \item "[o] with marks"
    \item "[o] with imperfections"
    \item "cracked [o]"
    \item "faulty [o]"
    \item "incomplete [o]"
    \item "bent [o]"
    \item "snapped [o]"
    \item "scratched [o]"
    \item "shattered [o]"
    \item "fractured [o]"
    \item "burst [o]"
    \item "[o] in pieces"
\end{itemize}
}
\end{minipage}
\begin{minipage}[t]{0.33\linewidth}
(b) Localising prompts

{\tt \small
\begin{itemize}
    \item "a tear"
    \item "a rip"
    \item "some damage"
    \item "a fault"
    \item "a break"
    \item "an abnormality"
    \item "a defect"
    \item "a crack"
    \item "an anomaly"
    \item "a missing component"
    \item "an error"
    \item "a mark"
    \item "a cut"
    \item "a dent"
    \item "a scratch"
    \item "an imperfection"
    \item "a blemish"
    \item "a mistake"
    \item "an error"
\end{itemize}
}
\end{minipage}
\caption{Lists of prompts used in our pipeline, excluding those from the original WinCLIP \cite{Jeong2023}.}
\label{fig:comp_prompt}
\end{figure*}
\end{document}